\pdfoutput=1

\documentclass[11pt]{article}

\usepackage[final]{acl}

\usepackage{times}
\usepackage{latexsym}

\usepackage[T1]{fontenc}

\usepackage[utf8]{inputenc}

\usepackage{microtype}

\usepackage{inconsolata}

\usepackage{graphicx}

\usepackage{booktabs}
\usepackage{multirow}
\usepackage{xcolor}

\usepackage{float}

\definecolor{customblue}{HTML}{2C6968}
\definecolor{custompink}{HTML}{B03A72}
\definecolor{bottlegreen}{rgb}{0.0, 0.416, 0.306}

\title{Breaking the Curse of Multilinguality with Cross-lingual \\ Expert Language Models}

\author{Terra Blevins$^{1\dagger}$ \quad  Tomasz Limisiewicz$^{2*}$ \quad Suchin Gururangan$^1$ \quad Margaret Li$^1$ \\ \bf Hila Gonen$^1$ \quad Noah A. Smith$^{1,3}$ \quad Luke Zettlemoyer$^1$ \\
$^1$Paul G. Allen School of Computer Science and Engineering, University of Washington \\ 
$^2$Faculty of Mathematics and Physics, Charles University in Prague \\
$^3$Allen Institute for Artificial Intelligence}

\begin{document}
\maketitle
\begingroup\def\thefootnote{$\dagger$}\footnotetext{Correspondence to {\tt blvns@cs.washington.edu}}\endgroup
\begingroup\def\thefootnote{*}\footnotetext{Work done while visiting the University of Washington.}\endgroup
\begin{abstract}
Despite their popularity in non-English NLP, multilingual language models often underperform monolingual ones due to inter-language competition for model parameters. We propose Cross-lingual Expert Language Models (\textsc{x-elm}), which mitigate this competition by independently training language models on subsets of the multilingual corpus. This process specializes \textsc{x-elm}s to different languages while remaining effective as a multilingual ensemble. Our experiments show that when given the same compute budget, \textsc{x-elm} outperforms jointly trained multilingual models across all 16 considered languages and that these gains transfer to downstream tasks. \textsc{x-elm} provides additional benefits over performance improvements: new experts can be iteratively added, adapting \textsc{x-elm} to new languages without catastrophic forgetting. Furthermore, training is asynchronous, reducing the hardware requirements for multilingual training and democratizing multilingual modeling.
\end{abstract}

\begin{figure*}[t]
    \centering
    \includegraphics[width=0.9\linewidth]{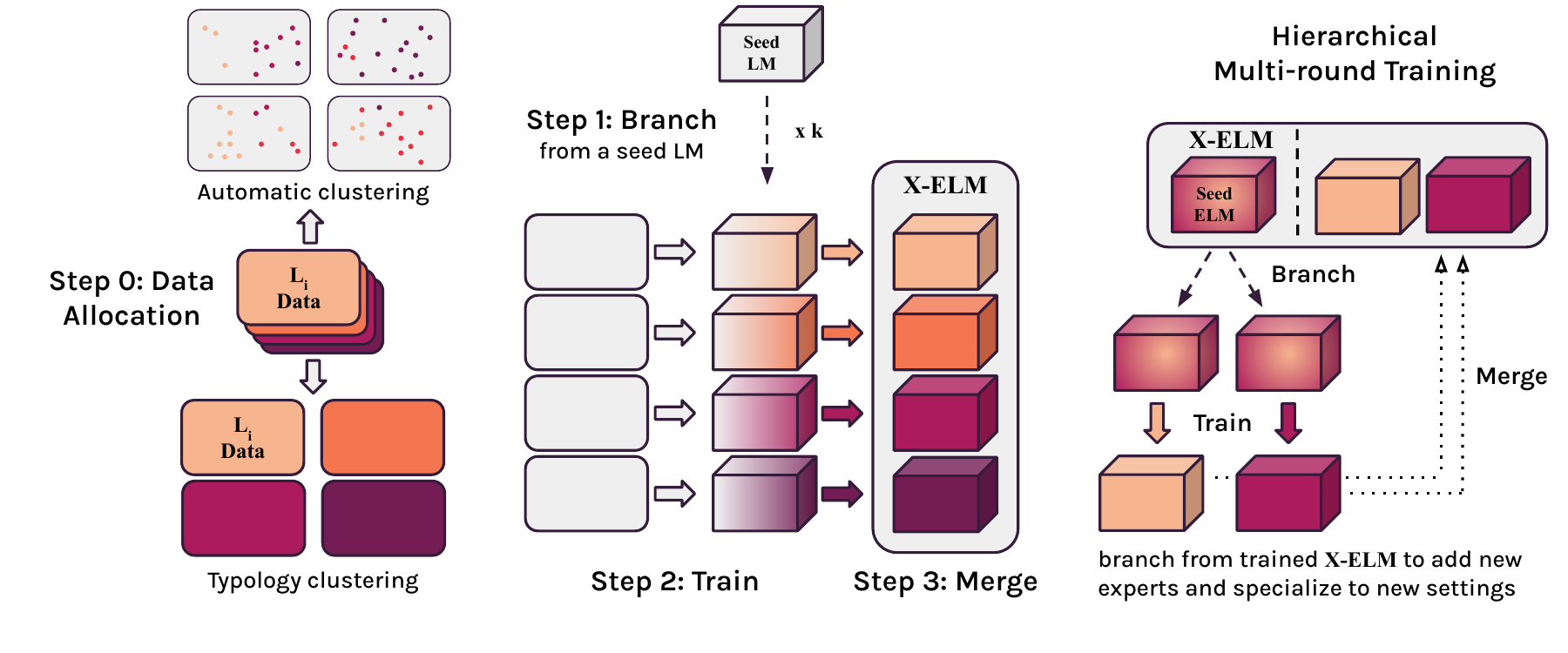}
    \caption{Overview of the \textsc{x-elm} pretraining procedure. \textbf{Left}: We partition the multilingual text corpus into $k$ subsets either through \textit{automatic TF-IDF} clustering of documents or through grouping languages by \textit{linguistic typology}. \textbf{Center}: Branch-Train-Merge (BTM) pretraining method. We initialize (\textit{branch}) $k$ experts from a seed LM, \textit{train} each expert on a different cluster from the pretraining corpus, and \textit{merge} the experts into a set of \textsc{x-elm}s. \textbf{Right}: Hierarchical Multi-Round (HMR) training procedure (\S \ref{sec:hmr-training}).}
    \label{fig:method}
\end{figure*}

\section{Introduction}

Massively multilingual language models (LMs), which are trained on terabytes of text in a hundred or more languages, underlie almost all non-English and cross-lingual NLP applications~\cite[i.a.]{workshop2022bloom, lin2022fewshot}. 
Despite their wide adoption, these models come at a cost: the many languages are represented in the same, fixed model capacity, causing performance on individual languages to degrade relative to monolingual models \cite{conneau2020unsupervised, chang2023multilinguality}. This phenomenon (termed the \textit{curse of multilinguality}) can significantly harm low-resource language performance \cite{wu2020languages}. 

In this paper, we address this \textit{curse} with \textbf{Cross}-lingual \textbf{E}xpert \textbf{L}anguage \textbf{M}odels (\textsc{x-elm}, Figure \ref{fig:method}), an ensemble of language models initialized from a pretrained multilingual model and each independently trained on a different portion of a multilingual corpus with x-BTM, a new extension of the Branch-Train-Merge paradigm \cite[BTM;][\S\ref{sec:background}]{li2022branchtrainmerge, gururangan2023scaling} to the more heterogenous multilingual setting. \textsc{x-elm} allows for efficient scaling of model capacity to better represent all considered languages.

x-BTM adapts existing BTM techniques to the multilingual setting by introducing a new cluster method for data assignments based on typological language similarity (\S \ref{sec:method-data}). We also propose Heirachical Multi-round Training (HMR; \S \ref{sec:hmr-training}), a method for efficiently adapting trained \textsc{x-elm}s to novel multilingual settings by branching from existing, typologically related \textsc{x-elm}s; this method for adapting \textsc{x-elm} to new languages strongly outperforms standard language adaptation methods.

We train \textsc{x-elm}s on 20 languages---including adapting to 4 unseen ones---with up to 21 billion training tokens, with the 1.7B parameter XGLM model as the base \cite{lin2022fewshot}. 
Our experiments show that \textsc{x-elm} strongly outperforms the dense multilingual models given the same compute budget in every considered setting (\S \ref{sec:results}), and that these improvements consistently benefit \textit{every} language. %
Notably, monolingual experts generally underperform typologically-informed multilingual \textsc{x-elm}s, indicating that linguistically targeted multilinguality can benefit language modeling.
We then show that the language modeling gains of \textsc{x-elm} hold on downstream evaluations (\S \ref{sec:icl-results}). 

Multilingual modeling with \textsc{x-elm} provides additional benefits beyond improved performance. Training a set of \textsc{x-elm}s is more computationally efficient than training a comparable dense model; each expert is trained independently, which removes the overhead cost of cross-GPU synchronization \cite{li2022branchtrainmerge} and allows asynchronous model training in low-compute settings. Similarly, adapting \textsc{x-elm}s to new languages with HMR training---a popular use case of language models \cite[i.e.,][]{chau2020parsing}---is more efficient than continued training of a dense LM and does not risk forgetting previously learned languages \cite{yogatama2019learning}.
As a result, \textsc{x-elm} allows much more efficient and effective multilingual modeling than prior approaches, democratizing multilingual NLP. We release the code and trained models.\footnote{\url{https://github.com/blvns/x-elm/}}

\section{Background: Branch-Train-Merge}
\label{sec:background}

Multilingual LMs are typically trained in a \textit{dense} manner, where a single set of parameters are updated with every training batch. When training large LMs, the dense training setup calculates gradients on and synchronizes model parameters across many GPUs.\footnote{For example, the XGLM-7.5B model ``was trained on 256 A100 GPUs for about 3 weeks'' \cite{lin2022fewshot}.} This requires all GPUs to be available simultaneously and incurs communication costs that prolong training.  

Branch-Train-Merge (BTM; \citealt{li2022branchtrainmerge}) alleviates this cost by dividing the total compute among smaller expert language models that are trained independently on different domains and then combined during inference time. While the total number of parameters increases with the number of experts, inference with these models often uses a subset of experts (see \S \ref{sec:method-inference}), keeping inference costs manageable. c-BTM \cite{gururangan2023scaling} then generalizes the above approach with automatic clustering of domains. Across multiple corpora, they show that (1) the optimal number of experts increases with data and compute and (2) a set of small expert models performs similarly to equivalently sized dense models at vastly reduced FLOP budgets.

Our work extends these studies to the multilingual setting, where experts are specialized to different languages instead of English-language domains. In the multilingual setting, we can also use typological structure to specialize experts, which we show provides additional benefits over automatic clustering. We also demonstrate that leveraging the hierarchy of language families in multi-round training yields further performance gains.

\section{Cross-lingual Expert Language Models}
\label{sec:method}

Multilingual language models are jointly trained on many different languages \cite[e.g.,][]{lin2022fewshot}, despite the well-documented effect this has on individual language performance \cite{conneau2020unsupervised, wang2020negative}.
We propose \textbf{Cross-lingual Expert Language Models}, or \textsc{x-elm}s (Figure \ref{fig:method}), which we hypothesize will alleviate the curse of multilinguality while maintaining the cross-lingual properties of dense multilingual LMs.

\subsection{x-BTM: Sparse Multilingual Training}
\label{sec:method-train}

This section overviews our algorithm for the sparse training of multilingual experts. 

\paragraph{Step 0: Multilingual Data Allocation} As a preprocessing step, we partition the multilingual corpus into $k$ clusters to train each \textsc{x-elm}. We consider both TF-IDF clusters and a new clustering method that groups documents by language identity and linguistic typology (\S \ref{sec:method-data}).

\paragraph{Step 1: Branch} A preliminary stage of shared, dense pretraining is important for ensembling expert language models \cite{li2022branchtrainmerge}. Therefore, the first step of BTM is to initialize (\textit{branch}) each expert with the parameters from a partially trained language model. For this work, we initialize our \textsc{x-elm}s with an existing multilingual pretrained model, XGLM \cite{lin2022fewshot}.

\paragraph{Step 2: Train} After initialization, we assign each expert a data cluster and train for a fixed number of steps with an autoregressive LM objective. Expert training is independent, with no shared parameters between models.

\paragraph{Step 3: Merge} We collect the $k$ \textsc{x-elm}s into a set and perform inference with them. We consider several methods of inference and expert ensembling in \S \ref{sec:method-inference}. \\

\noindent Steps 1--3 describe a single round of x-BTM training. However, we can continue to update the \textsc{x-elm} set by branching---initializing a new group of experts---from existing models in the ensemble and performing more rounds of x-BTM via the method proposed in \S \ref{sec:hmr-training}. This allows us to further improve \textsc{x-ELM} by training and adding new experts.

\subsection{Data Allocation Methods}
\label{sec:method-data}
How we assign data to experts is a key component of training \textsc{x-elm}, particularly as the data becomes more diverse (i.e., spanning many languages). We consider two methods of data allocation: 

\paragraph{Balanced TF-IDF Clustering} We partition the multilingual corpus automatically into $k$ components with $k$-means clustering. First, we encode each document into a word-level TF-IDF representation;\footnote{Data tokenization is independent of the downstream model. Here, we use the sklearn text-vectorizer tokenizer.} we then perform balanced $k$-means clustering on these representations to obtain approximately balanced subsets of the data on which to train each \textsc{x-elm}. Further details on the balanced $k$-means clustering method can be found in \citet{gururangan2023scaling}. This allocation method uses no language information outside of what is inherent in the text (e.g., script, vocabulary).

\begin{figure}[h!]
    \centering
    \includegraphics[width=\linewidth]{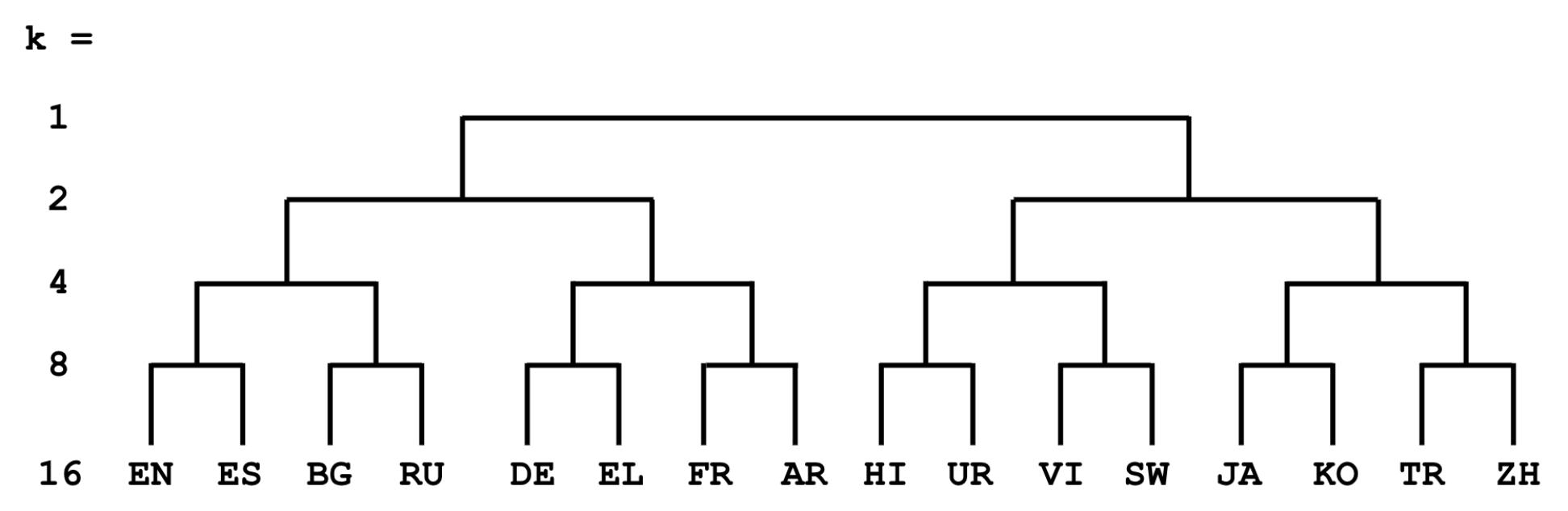}
    \caption{Hierarchical clustering of languages used to train our \textsc{x-elm} ensembles.} %
    \vspace{-5pt}
    \label{fig:lang-tree}
\end{figure}

\paragraph{Linguistic Typology Clustering} We also consider segmenting the corpus by language identity.\footnote{This requires knowledge of each document's language. We use the language tags provided with mC4.} Rather than balancing the amount of data allocated to each cluster in this setting, we keep the number of languages per cluster fixed. Specifically, we learn a balanced hierarchical clustering of the languages (Figure \ref{fig:lang-tree}). We build this hierarchy using the language similarity metrics in \textsc{lang2vec} \cite{littell2017uriel}, which represents languages based on linguistic features in resources such as WALS\footnote{World Atlas of Language Structures, \url{https://wals.info/}} and estimates language similarity with distance in this feature space. We initialize each cluster with a single language; at each step, we merge each cluster with exactly one other based on the \textit{minimum} distances between cluster centroids. We then group languages by the resulting hierarchy and desired number of experts. When the number of languages equals the number of experts, typological clustering results in monolingual training, as every language is assigned a separate expert.

\subsection{Inference with \textsc{x-elm}s}
\label{sec:method-inference}
We evaluate multiple methods for performing inference with \textsc{x-elm}s:

\paragraph{Top-1 Expert} This method performs inference with a single expert chosen prior to evaluation, which incurs the same inference cost as the dense baselines. When evaluating \textit{Typology} experts on a particular language $\ell$, we choose the expert that included $\ell$ in the set of languages on which they continued pretraining. Similarly, when evaluating \textit{TF-IDF}, we choose the \textsc{x-elm} trained on the highest percentage of $\ell$'s data.

\paragraph{Ensembling TF-IDF Experts} We also consider ensembling TF-IDF experts by adapting the c-BTM routing method \cite{gururangan2023scaling}. Here, we calculate ensembling \textbf{$\alpha$}, or weights, over the experts at \textit{each} inference step based on the prior context's TF-IDF distance from the experts' $k$-means centroids. These weights are then used to ensemble the output probabilities from each expert. 

More specifically, given a probability from each expert $p_{e}(x_t\mid x_{<t})$ and the corresponding ensemble weight $\alpha_e = p(e\mid x_{<t}) \propto \textrm{exp} (- \textrm{dist}(x_{<t},c_{e})^2/T)$, the probability of the ensemble is $$p_{E}(x_t\mid x_{<t}) = \sum_{e \in E} \alpha_e \cdot p_{e}(x_t\mid x_{<t})$$ Here, $\textrm{dist}(x_{<t},c_{e})$ is obtained by embedding $x_{<t}$ with the learned TF-IDF vectorizer and calculating the Euclidean distance from $c_e$ (the centroid over the data representations allocated to expert $e$), and $T$ is a temperature parameter over the ensemble weight distribution. Further details and motivation are provided in \citet{gururangan2023scaling}.

Ensembling \textsc{x-elm} outputs increases the cost of inference relative to dense models or top-1 inference. However, it can potentially better fit different subsets of data in a diverse evaluation set. We also do not assume we know the language of each example when ensembling, which makes this approach more flexible than the top-1 setting.
In most cases, we ensemble all $k$ experts; however, we can also reduce computational costs by \textit{sparsifying} the ensemble and only activating the $m$ ($<k$) experts that most contribute to an example:  $p_{E}(x_t\mid x_{<t}) = \sum_{e \in E} \alpha_e \cdot p_{e}(x_t\mid x_{<t}) : \alpha_e \in \textrm{top-m}(\bf{\alpha}_E)$. Appendix Table \ref{tab:app-sparse-tfidf} presents the performance tradeoff with sparser TF-IDF ensembles.

\begin{figure*}
    \centering
    \includegraphics[width=\linewidth]{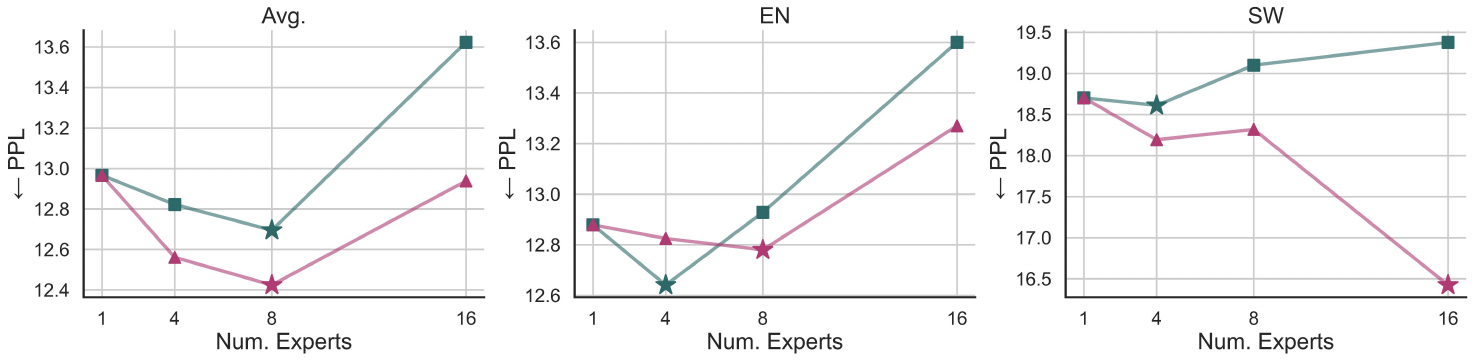}
    \caption{Average and language-specific (EN and SW) perplexities across different expert counts $k$ (Num. Experts). In each evaluation setting, we compare clustering training data for experts with the \textcolor{customblue}{TF-IDF$_{top1}$} (\textcolor{customblue}{square}) and \textcolor{custompink}{Linguistic Typology} (\textcolor{custompink}{triangle}) methods (\S \ref{sec:method-data}). The best choice of $k$ for each setting is marked with a star.}
    \label{fig:num-expert-comp}
\end{figure*}

\section{Hierarchical Multi-Round Training}
\label{sec:hmr-training}

We previously described a single round of training for \textsc{x-elm} (\S \ref{sec:method-train}). However, BTM can also be used iteratively to train new experts seeded with those learned in a prior round. The multilingual setting provides a natural extension of multi-round training that leverages typological structure. %

We propose \textbf{Hierarchical Multi-Round} (\textbf{HMR}) pretraining (Figure \ref{fig:method}), which uses the learned typological tree structure from \textit{Linguistic Typology} clustering to iteratively train more specific \textsc{x-elm}s. Specifically, given an expert model $x$  trained on a cluster of languages $L$, we initialize a new set of experts $X' = x_1', x_2',...,x_n'$ with the parent expert $x$. Each new expert in $X'$ is then further trained on a different sub-cluster $\ell' \subset L$. 

HMR pretraining gives multiple benefits over single-round BTM. In particular, HMR training saves compute and more easily adapts our \textsc{x-elm}s to new settings. A specific application of this is adding new languages to the model: while updating dense multilingual LMs with new languages is difficult and can lead to catastrophic forgetting of existing languages \cite{winata2023overcoming}, hierarchically training an expert on a new language adds it to the \textsc{x-elm} set without altering the existing information in other experts. We evaluate HMR training with this use case in \S \ref{sec:results-unseen}.

\section{Experimental Design}

\subsection{Pretraining Data and Languages}
We train our \textsc{x-elm}s on mC4, an open-source, multilingual pretraining corpus derived from CommonCrawl \cite{xue2021mt5}.\footnote{While one could also continue pretraining with the same corpus that the seed LM was trained on, the pretraining data for XGLM is not publicly available.} mC4 provides language tags for each document in the corpus, which were automatically assigned with cld3\footnote{\url{https://github.com/google/cld3}} when the dataset was constructed; we use these language tags during typological clustering (\S \ref{sec:method-data}). We focus our experiments on the 16 highest-resourced languages out of the 30 languages on which the seed LM, XGLM-1.7B, was trained. For languages with significantly more data than the others (e.g., English), we subsample their data to the first 1,024 shards. Appendix Table \ref{tab:app-data} gives the languages and data quantities in our pretraining corpus.

\subsection{Pretraining Settings}
Each expert in the \textsc{x-elm} experiments is a 1.7B parameter model with the same architecture as the 1.7B XGLM transformer model \cite{lin2022fewshot}, and they are initialized with XGLM's weights in the initial round of BTM training. Unless otherwise stated, we keep the training parameters from the original XGLM training procedure; further details are given in Appendix \ref{app:params}.

We train the experts for a fixed number of training steps. The exact parameters and resources used for each \textsc{x-elm} experiment are reported in Table \ref{tab:app-resources}: in every setting, we control for the number of tokens seen during training. This ensures that all experts in a setting see the same amount of data (and undergo the same number of training updates) and that experiments across different expert set sizes but under the same training budget are comparable. For most experiments, we use a shared budget of 10.5B tokens; where indicated, we increase this to 21.0B tokens to test the effect of further training.

\subsection{Baselines} 
We compare the performance of our \textsc{x-elm} experiments to two dense baselines: \textbf{XGLM}, which is the 1.7B parameter seed model used to initialize each expert prior to x-BTM training, and a single \textbf{dense} model, which is also initialized with XGLM-1.7B's weights and then trained with the full data and compute budget split across experts in other  \textsc{x-elm} settings. This is equivalent to $k$=1 experts in cases where we vary the number of experts. 

Since the curse of multilinguality is often evaluated in comparison to monolingual modeling, We also consider the setting where we train \textbf{monolingual} expert models on each target language (\S \ref{sec:results-choose-k}). Given that we consider sixteen languages in our \textsc{x-elm} experiments, this corresponds to the $k=16$ typological clustering setting.

\subsection{Perplexity Evaluation}
To evaluate the language modeling performance of the \textsc{x-elm}s, we separately calculate the perplexity on the mC4 validation sets of each pretraining language. For languages with larger evaluation sets, we estimate performance on the first 5,000 validation examples. This perplexity metric is not comparable across languages, as they have different validation sets.

\section{Language Modeling Experiments}
\label{sec:results}
We now test the effectiveness of sparse language modeling in the multilingual setting. First, we determine the optimal number of clusters for our given compute budget and dataset (\S \ref{sec:results-choose-k}). We then demonstrate that \textsc{x-elm}s outperform comparable dense models on seen languages (\S \ref{sec:seen-results}) and more effectively adapt to new, unseen languages (\S \ref{sec:results-unseen}).

\subsection{Choosing the Number of \textsc{x-elm}s}
\label{sec:results-choose-k}
We first consider which number of experts gives the best multilingual language modeling performance. Figure \ref{fig:num-expert-comp} compares the choice of \mbox{$k\in \{1, 4, 8, 16\}$} \textsc{x-elm}s when trained on 10.5B tokens.\footnote{The $k=16$ setting is equivalent to training monolingual experts for every language. Full results are in Table \ref{tab:ppl-results} for $k=8$ and Appendix Table \ref{tab:app-ppl-results} for $k=4$ and $k=16$.} $k=8$ is the best-performing setting on 75\% of languages when clustering with TF-IDF and for 15 of the 16 pretraining languages when clustering by language similarity. Furthermore, typological clustering consistently outperforms TF-IDF.

These experiments indicate that, for the budget we evaluate, \textbf{the best overall \textsc{x-elm} setting is bilingual models (k=8) clustered by language similarity}. This result is surprising, as it is intuitive to assume that simply continuing to pretrain each expert on a single language (i.e., the $k=16$ setting) would lead to better perplexity. We find that one language, Swahili, does benefit from the monolingual $k=16$ setting---possibly because Swahili is paired with a distant language (Vietnamese) by the typological clustering process. However, perplexity is higher in the $k=16$ setting for all other languages, and in some cases, even underperforms the dense ($k=1$) model. 

\begin{table*}[]
    \centering
    \small
    \setlength{\tabcolsep}{5 pt}{
    \begin{tabular}{c | r | r r r r | r r r r }
        \toprule
        \multirow{2}{*}{\textbf{Lang.}} & & \multicolumn{4}{c}{\textbf{10.5B Training Tokens}} & \multicolumn{4}{|c}{\textbf{21.0B Training Tokens}} \\
        & XGLM & Dense & TF-IDF$_{top1}$ & TF-IDF$_{ens}$$^*$ & Typ. & Dense & TF-IDF$_{top1}$ & TF-IDF$_{ens}$$^*$ & Typ. \\
        \addlinespace
        \textbf{AR} & 16.85 & 15.29 & \textbf{14.51} & 14.56 & 14.66 & 14.97 & \textbf{14.00} & 14.05 & 14.16 \\
        \textbf{BG} & 11.31 & 10.44 & 10.39 & 10.39 & \textbf{10.25} & 10.34 & 10.27 & 10.26 & \textbf{10.09} \\
        \textbf{DE} & 15.53 & 14.02 & \textbf{13.41} & 13.50 & 13.42 & 13.72 & \textbf{12.95} & 13.05 & 12.97 \\
        \textbf{EL} & 10.44 & 9.40 & 9.20 & 9.18 & \textbf{9.17} & 9.24 & 9.03 & 9.00 & \textbf{8.98} \\
        \textbf{EN} & 14.37 & 12.88 & 12.93 & \textbf{12.73} & 12.78 & 12.69 & 12.68 & \textbf{12.47} & 12.55 \\
        \textbf{ES} & 16.02 & 14.13 & 13.92 & \textbf{13.76} & 13.99 & 13.87 & 13.54 & \textbf{13.37} & 13.69 \\
        \textbf{FR} & 13.12 & 11.78 & \textbf{11.19} & 11.28 & 11.29 & 11.54 & \textbf{10.79} & 10.88 & 10.91 \\
        \textbf{HI} & 18.28 & 14.28 & 14.86 & 14.19 & \textbf{11.25} & 13.68 & 14.36 & 13.62 & \textbf{10.52} \\
        \textbf{JA} & 14.57 & 12.31 & 11.95 & 11.95 & \textbf{11.49} & 11.79 & 11.36 & 11.37 & \textbf{10.88} \\
        \textbf{KO} & 8.82 & 7.79 & 7.72 & \textbf{7.67} & \textbf{7.67} & 7.67 & 7.61 & \textbf{7.53} & 7.54 \\
        \textbf{RU} & 13.43 & 12.52 & 12.14 & 12.21 & \textbf{12.08} & 12.33 & 11.83 & 11.90 & \textbf{11.74} \\
        \textbf{SW} & 19.85 & 18.70 & 19.10 & 18.76 & \textbf{18.32} & 18.61 & 19.04 & 18.67 & \textbf{18.07} \\
        \textbf{TR} & 17.81 & 15.34 & 14.13 & 14.28 & \textbf{13.80} & 14.88 & 13.41 & 13.58 & \textbf{13.03} \\
        \textbf{UR} & 14.38 & 13.45 & 13.40 & 13.57 & \textbf{12.60} & 13.38 & 13.26 & 13.52 & \textbf{12.20} \\
        \textbf{VI} & 13.07 & 11.39 & 11.00 & 10.86 & \textbf{10.22} & 11.09 & 10.56 & 10.42 & \textbf{9.69} \\
        \textbf{ZH} & 17.91 & 13.74 & 13.28 & 13.53 & \textbf{11.98} & 13.12 & 12.61 & 12.87 & \textbf{11.24} \\
        \addlinespace
        \textbf{Avg.} & 14.74 & 12.97 & 12.70 & 12.60 & \textbf{12.19} & 12.68 & 12.33 & 12.28 & \textbf{11.77} \\
        \bottomrule
    \end{tabular} }
    \caption{Per-language and average perplexity results for the $k=8$ \textsc{x-elm} experiments (original XGLM and $k=1$ dense model included for comparison). Lower numbers are better. The best setting for each language is bolded per compute budget. $^*$TF-IDF ensemble uses more parameters for inference than other evaluations; see Table \ref{tab:app-sparse-tfidf} for the effect of sparsifying these ensembles on perplexity.}
    \label{tab:ppl-results}
\end{table*}

\subsection{Perplexity Results on Seen Languages}
\label{sec:seen-results}
We now examine the performance of \textsc{x-elm} in the best setting ($k$ = 8) for the sixteen languages seen during BTM training on computational budgets of 10.5B and 21.0B tokens. Table \ref{tab:ppl-results} presents the perplexities of the TF-IDF clustered \textsc{x-elm}s as well as the typologically (Typ.) clustered \textsc{x-elm}s. As baselines, we compare against the original XGLM-1.7B model and a dense model trained on both computational budgets. We find that the best setting, $k=8$ with typologically clustered experts, improves by 2.97 and 1.20 on average over the seed and dense baseline models and has individual language gains of up to 7.77 and 3.76 over these models, respectively.

\paragraph{Expert language models outperform dense continued training} For most languages (10 of 16), typologically clustered experts are the best-performing setting. For some high-resource languages (EN and ES), ensembling the TF-IDF experts works better than a single expert. However, this inference setting requires more parameters, as it uses all \textsc{x-elm}s instead of just the single best expert per language. Furthermore, training \textsc{x-elm}s for longer unsurprisingly outperforms lower compute settings.
All of our experimental settings outperform the seed XGLM model; similarly, the experiments with the 21.0B token compute budget perform better than the respective experiment trained with 10.5B tokens.

\begin{figure}[b!]
    \centering
    \includegraphics[width=\linewidth]{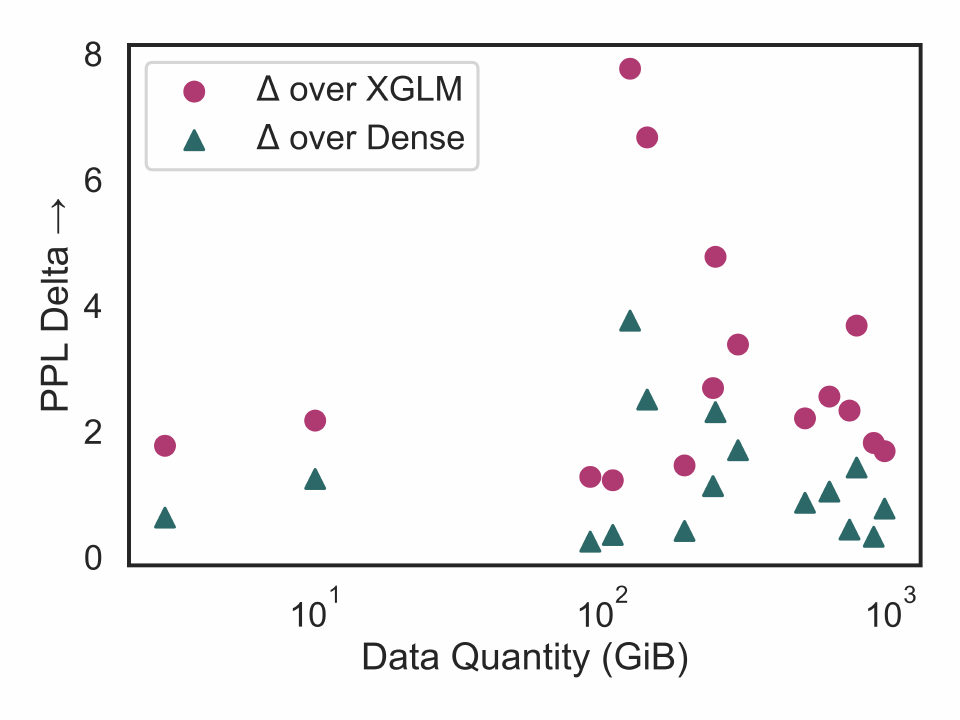}
    \caption{PPL improvements per language over XGLM-1.7B (\textcolor{custompink}{circle}) and dense baseline (\textcolor{customblue}{triangle}) against the training data quantity (for typ. clustered experts). }
    \label{fig:ppl-by-data}
\end{figure}

\paragraph{\textsc{x-elm}s improve language modeling on all languages} We also show that multilingual language modeling with \textsc{x-elm}s does not disproportionally benefit languages with more pretraining data (Figure \ref{fig:ppl-by-data}). Instead, perplexity improvements over both the seed LM and the dense LM baseline \textit{may} slightly favor low-resource languages ($\rho=-0.19, -0.26$, respectively).

\subsection{Unseen Languages and Modeling New Languages with \textsc{x-elm}}
\label{sec:results-unseen}
A common method for applying multilingual language models to new settings is language-adaptive pretraining \cite[LAPT;][]{chau2020parsing}), as this reuses multilingual knowledge in existing LMs while training on the target language(s). We now examine how well \textsc{x-elm} performs in this paradigm by (1) evaluating \textsc{x-elm} perplexity on unseen languages and (2) adapting an existing \textsc{x-elm} set to new languages. Specifically, we consider both zero-shot evaluation and further training of \textsc{x-elm} on four languages not included in the original XGLM seed model: Azerbaijani (AZ), Hebrew (HE), Polish (PL), and Swedish (SV).\footnote{Data for these languages is also obtained from mC4, with the same preprocessing as other languages in our experiments.} 

\paragraph{Unseen Language Evaluation} We evaluate the existing dense baseline and ensembled TF-IDF clustered experts from the 21B token compute budget (\S \ref{sec:seen-results}) to test whether continued pretraining with x-BTM improves performance on unseen languages (\textbf{\textsc{x-elm} Training}). We also compare these results to XGLM. We note these models \textit{never} trained on the target languages.

Table \ref{tab:unseen-ppl} presents the unseen target language perplexities in the \textbf{XGLM} and \textbf{\textsc{x-elm} Training} columns. We find that the original XGLM model performs poorly on the new languages, particularly those less related to XGLM's highest-resourced ones (i.e., AZ and HE). While the perplexities remain high in the dense model and TF-IDF ensembles, continued training on other languages improves the seed model.

\begin{table}[b!]
    \footnotesize %
    \centering
    \setlength{\tabcolsep}{5 pt}{
    \resizebox{0.99\columnwidth}{!}{\begin{tabular}{c | r | r r | r r}
        \toprule
        \multirow{2}{*}{\textbf{Lang}} & & \multicolumn{2}{c|}{\textbf{\textsc{x-elm} Training}}  & \multicolumn{2}{c}{\textbf{LAPT}} \\
         & XGLM & Dense & TF-IDF$_{ens}^{*}$ & Dense & HMR \\
         \multicolumn{1}{c}{\textbf{Target}} \\
         AZ & 1467.45 & 739.58 & 722.10 & 65.73 & \textbf{32.74} \\
         HE & 1817.07 & 685.02 & 815.96 & 53.08 & \textbf{26.21} \\
         PL & 211.76 & 160.70 & 178.63 & 17.71 & \textbf{16.60} \\
         SV & 105.27 & 92.55 & 99.24 & 27.37 & \textbf{26.16} \\   
        \multicolumn{1}{c}{\textbf{Donor}} \\
        TR & 17.81 & 15.34 & 14.28 & 14.69 & \textbf{12.72} \\
        AR & 16.85 & 15.29 & 14.56 & 14.80 & \textbf{13.52} \\
        RU & 13.43 & 12.52 & 12.21 & 12.28 & 12.02 \\
        EN & 14.37 & 12.88 & 12.73 & 12.65 & 12.63 \\
        \bottomrule
         
    \end{tabular} }
    }
    \caption{Perplexity results on unseen target languages and their respective donor languages. Donor language performance is only \textbf{bolded} if these results outperform all other \textsc{x-elm} settings in that language (Table \ref{tab:ppl-results}).}
    \label{tab:unseen-ppl}
\end{table}

\begin{table*}[]
    \centering
    \small
    \begin{tabular}{l l| r r r r r r}
    \toprule
      & \multirow{2}{*}{\textbf{Model}} & \multicolumn{2}{c}{\textbf{XNLI}} & \multicolumn{2}{c}{\textbf{XStoryCloze}} & \multicolumn{2}{c}{\textbf{PAWS-X}} \\
       && Acc. & Win Rate & Acc. & Win Rate & Acc. & Win Rate \\
       \addlinespace
       \multirow{6}{*}{\textbf{Zero-shot}} & XGLM (1.7B) & 44.88 & 28.6\% & 57.76 & \textbf{28.6\%} & 48.54 & 14.3\% \\ %
       & Dense & 44.31 & 7.1\% & 56.10 & 0.0\% & 48.44 & 28.6\% \\ %
       & Typ. (TRG) & 44.17 & 7.1\% & 57.79 & \textbf{28.6\%} & 49.86 & \textbf{42.9\%} \\ %
       & TF-IDF (Top-1) & 43.77 & 14.3\% & \textbf{57.80} & \textbf{28.6\%} & \textbf{50.04} & 28.6\% \\ %
       & TF-IDF (Ens.) & \textbf{45.10} & \textbf{42.9\%} & 57.46 & 14.3\% & 49.93 & 0.0\% \\ %
       \addlinespace
       \multirow{6}{*}{\textbf{Few-shot}} & XGLM (1.7B) & 42.34 & 28.6\% & 53.21 & 0.0\% & 54.52 & 0.0\% \\
       & Dense & 41.70 & 0.0\% & 55.00 & 0.0\% & 54.81 & 14.3\% \\
       & Typ. (TRG) & 42.15 & $^\dagger$14.3\% & 54.62 & $^\dagger$\textbf{71.4\%} & 55.39 & $^\dagger$28.6\% \\
       & Typ. (EN) & 42.43 & $^\dagger$7.1\% & \textbf{55.54} & $^\dagger$28.6\% & 55.13 & 14.3\% \\
       & TF-IDF (Top-1) & 42.55 & 21.4\% & 55.03 & $^\dagger$14.3\% & \textbf{55.50} & $^\dagger$\textbf{42.9\%} \\
       & TF-IDF (Ens.) & \textbf{42.93} & \textbf{35.7\%} & 54.72 & 28.6\% & 54.57 & 14.3\% \\
    \bottomrule
    \end{tabular}
    \caption{Average performance (\textbf{Acc.}) and the percentage of languages where this setting outperforms the others (\textbf{Win Rate}) on the overlap of task evaluation languages and the \textsc{x-elm} target languages. The \textbf{few-shot} setting provides $k$=8 English demonstrations to the model and averages performance across five runs. $^\dagger$Indicates (best) performance ties between two evaluation settings on a language when calculating the win rate.}
    \label{tab:downstream-summ}
\end{table*}

\paragraph{Adapting \textsc{x-elm} to new languages} We now consider how well Hierarchical Multi-Round training (\textbf{HMR}, \S \ref{sec:hmr-training}) works for language adaptive pretraining \cite[LAPT,][]{chau2020parsing}, which incorporates new target languages into the continued pretraining process. Here, we group each \textit{target} language with a higher-resource \textit{donor} language already in our pretraining set; these are assigned with the language similarity metric used for typological clustering. We seed each new language's expert with an expert specialized to that language's donor; the new expert is then trained on the donor/target language pair. For HMR inference, we evaluate perplexity with the expert trained for that target language; we also evaluate on the donor languages to see what benefit, if any, they receive from the adaptation process.

We compare HMR against jointly continuing training on all four new languages and their respective donors in a single model (\textbf{Dense}). Each setting builds on models from the 10.5B compute budget: we continue training on the dense baseline for dense LAPT and branch from the donor languages' $k$=8 typological experts for HMR training.

All of the LAPT settings provide considerable improvements on the new target languages over the unseen language experiments (Table \ref{tab:unseen-ppl}, \textbf{LAPT} columns). The HMR setting outperforms continued dense training on every new language. Furthermore, HMR training removes the risk of \textit{catastrophic forgetting} \cite{yogatama2019learning} in other LAPT schemes, as this process adds new experts to \textsc{x-elm} rather than changing existing ones.\footnote{This forgetting of known languages occurs in our dense LAPT baseline, with perplexity decreasing by 1.91 points on average for languages not included in the adaptation setting.}

We also find that this setting provides performance gains on two donor languages over the experiments in \S \ref{sec:seen-results}. This is likely due to further training with more closely related languages for these languages (e.g., performing training on Arabic with Hebrew rather than French), consequently providing a more informative training signal for the higher-resource donor language as well.

\section{In-Context Learning Experiments}
\label{sec:icl-results}
We also measure whether the perplexity improvements in \textsc{x-elm}s correspond to better performance on downstream tasks. We test \textsc{x-elm}s on three tasks in an in-context learning (ICL) framework, showing \textsc{x-elm} language modeling gains \textit{do} translate to ICL improvements over the baseline models.

\subsection{Experimental Setup}
We test the in-context learning abilities of \textsc{x-elm} on three downstream tasks:

\textbf{XNLI} \cite{conneau2018xnli} is a multilingual natural language inference benchmark covering 14 of our 16 pretraining languages (excluding JA and KO). Since there are no gold training examples for XNLI, we use the test set for evaluation and sample demonstrations from the validation set.

\textbf{XStoryCloze} \cite{lin2022fewshot} is a manually translated benchmark extending StoryCloze \cite{mostafazadeh2016corpus} to other languages. This is a story-completion task wherein the model identifies the correct final sentence of a short story. This dataset covers seven of our pretraining languages and four other low-resource languages. 

\textbf{PAWS-X} \cite{yang2019paws} is a binary classification task that requires the model to determine whether a pair of sentences are paraphrases. This benchmark covers seven of our pretraining languages, including two (JA and KO) that are not covered by the other ICL benchmarks. \\

We compare \textsc{x-elm}s against dense baselines in zero- and few-shot settings. For all benchmarks, we evaluate on 1,000 random examples and perform five runs on different demonstrations for few-shot learning, using English demonstrations for every language to test cross-lingual transfer.
Further details about the ICL evaluation are in Appendix \ref{app:icl}.

\subsection{Results}
We evaluate our best \textsc{x-elm}s by perplexity---$k$=8 experts trained on the larger compute budget of 21B training tokens---on their in-context learning abilities (Table \ref{tab:downstream-summ}).\footnote{Full results for each task are given in Appendix Tables \ref{tab:downstream-xnli}, \ref{tab:downstream-xstorycloze}, and \ref{tab:downstream-pawsx}.} Here, we consider two metrics to summarize model performance across languages: \textbf{Acc.} is the average accuracy of each model for that ICL task across evaluation languages, and \textbf{Win Rate} is the percentage of languages where that model achieves the highest score out of the considered models; if two models get the highest score, they are both considered to \textit{win} that setting.

We find that \textsc{x-elm}s outperform both the seed and compute-matched dense baselines across the three tasks and in both the zero- and few-shot evaluation settings. Furthermore, though \textsc{x-elm} improves over the seed model, the dense model underperforms XGLM. This may be due to using different data from the original XGLM pretraining, as data quality issues have been documented for mC4 \cite{kreutzer2022quality, chung2023unimax}. We also note that few-shot ICL performance on XNLI and XStoryCloze is consistently lower than in the zero-shot setting; this is a recurring issue in multilingual ICL also observed in the seed model \cite{lin2022fewshot}.

\section{Related Work}

\subsection{Multilingual Pretraining and Adapation}
Many variants of dense multilingual pretraining have been proposed since multilingual BERT \cite{bert}: changing the architecture and scaling the model size up \cite{xlmr, lin2022fewshot}, adding additional cross-lingual objectives \cite{xlm,xlme,paradise}, and careful language and data curation \cite{workshop2022bloom}.
Most similar to our work is \citet{pfeiffer2022lifting}, which proposes an architecture, X-MOD, with language-specific modules. However, many slimitations of dense modeling persist here as the model and language modules are jointly trained.

Across most multilingual pretraining methods is the \textit{curse of multilinguality} \cite{conneau2020unsupervised}, particularly for lower-resource languages in massively multilingual training \cite{wu2020languages}.
\newcite{blevins-etal-2022-analyzing} find that multilingual models forget information previously learned during training, which they hypothesize is due to this phenomenon; \citet{wang2020negative} similarly suggest that this effect is due to training dynamics. More recently, \citet{chang2023multilinguality} presented a controlled study corroborating limited model capacity as a cause of this  \textit{curse}. A primary motivation of our work is to mitigate this \textit{curse} while maintaining the other benefits of multilingual modeling.

However, not all multilinguality is harmful to language modeling. \citet{chang2023multilinguality} show that seeing linguistically similar languages can benefit low-resource language performance; this corroborates our finding that \textsc{x-elms} trained on related languages outperfrom monolingual experts. In this vein, a recent direction in multilinguality has been \textit{targeted multilingual modeling}, where models are trained on data from the same language family \cite{ogueji2021small, ogunremi2023mini, ljubevsic2024language, downey2024targeted}.

We also consider how \textsc{x-elm} can be used for language adaption. The most common method, language-adaptive pretraining \cite[LAPT;][]{chau2020parsing}, continues multilingual pretraining with new languages incorporated into the training regime. Other work proposed using adapters to update the model with new languages \cite{pfeiffer2020madx}; notably, \citet{faisal2022phylogeny} used similar linguistic motivations to our typological clustering to group languages for adapters. However, \citet{ebrahimi2021adapt} found that LAPT outperformed adapters for adaptation.

\subsection{Sparse Models for NLP}
Sparse language models \citep{pmlr-v119-evci20a,pmlr-v97-mostafa19a,dettmers-sparse-from-scratch} route inputs through a subset of the total model parameters. Our work builds most directly on the Branch-Train-Merge \citep{li2022branchtrainmerge, gururangan2023scaling} algorithm, which results in full-model experts specialized on domains defined by metadata or a learned clustering. This design expands on early Mixture-of-Experts (MoE) models \citep{jacobs1991adaptive} and on DEMix layers \citep{demix}, which routes sequences to per-layer experts based on metadata. %

Other MoE models have recently been applied to multilingual settings. \citet{pfeiffer2022lifting} develop a multilingual expert model with language-specific routing, and \citet{kudugunta-etal-2021-beyond-distillation} develop a machine translation model with routing determined by the source-target language pair or the target language. Similar to BTM, \citet{Jang2023ExploringTB} trains experts specialized to different tasks, including five machine-translation language pairs. %

\section{Conclusion}
This work presents an approach to break the \textit{curse of multilinguality} by extending sparse language modeling to the multilingual setting with \textsc{x-elm}. We find that \textsc{x-elm}s achieve better perplexity \textit{on every language} over standard, dense language models trained with the same compute budget; expert language models can also be easily adapted to new languages without catastrophic forgetting. \textsc{x-elm}s present additional benefits over dense models for multilingual modeling, including training efficiency and flexibility. Finally, we show that these language modeling improvements transfer to downstream, in-context learning performance.

While our experiments show that \textsc{x-elm} outperforms dense LMs, we foresee many avenues of future work to further tailor sparse modeling to multilinguality. These include better methods for data allocation---such as clustering methods that leverage cross-lingual signal--- and algorithmic improvements to better allocate compute and more effectively ensemble models. By proving the efficacy of sparse language modeling in the multilingual setting, we hope to inspire future work in this vein that fairly models every language while leveraging the potential of cross-lingual learning.

\section*{Limitations}
This work focuses on rigorously examining the effect of training \textsc{x-elm}s in a limited number of settings, training languages, and data sources; this is both to ensure that we provide comprehensive comparisons with prior approaches to multilingual language modeling and due to computational limitations. Therefore, the proposed method should be further verified in other settings. In particular, in future work, we hope to examine how \textsc{x-elm} performs at scale when using larger experts, more languages, and larger training budgets. Additionally, while we consider one seed model, XGLM \cite{lin2022fewshot}, future work should examine the effect of other pretrained initializations as well as training our own seed models to test how much multilinguality is needed in \textsc{x-elm} initialization.

We also note the limited nature of our downstream evaluations, which is due to (1) the limited number of multilingual benchmarks available and (2) our requirement that evaluation benchmarks overlap with (most of) our 16 pretraining languages. Furthermore, since we compare against the seed model, we focus on XGLM's original evaluation tasks and the prompting settings developed for this baseline (rather than developing our own that may be biased towards the \textsc{x-elm} models). 

Finally, training \textsc{x-elm} rather than a single dense model increases some computational costs, similar to other BTM methods. The primary increase is in storage, as each expert's weights need to be stored separately. In some cases, the inference cost of \textsc{x-elm} can be higher than the best model (e.g., when using an ensemble of experts); however, we propose several inference methods that only require loading a single model and demonstrate that you can sparsify the TF-IDF ensemble and achieve similar perplexities (Appendix Table \ref{tab:app-sparse-tfidf}).

\section*{Acknowledgements}
We would like to thank Orevaoghene Ahia for helpful feedback on this project. Tomasz Limisiewicz acknowledges the support of grant 338521 of the Charles University Grant Agency, Fellowship from Paul G. Allen School, and the Mobility Fund of Charles University.

\bibliography{custom}
\bibliographystyle{acl_natbib}
\appendix

\begin{table}[b!]
    \centering
    \small
    \resizebox{0.99\linewidth}{!}{ \begin{tabular}{r | r r r r} 
        \toprule
        \textbf{\# Tokens} & \textbf{k} & \textbf{\# GPUs} & \textbf{\# updates} & \textbf{grad acc.} \\
        \multirow[c]{4}{*}{\textbf{10.5 B}}  & 1 & 8 & 20,000 & 32 \\
        & 4 & 4 & 20,000 & 16 \\
        & 8 & 4 & 20,000 & 8 \\
        & 16 & 2 & 20,000 & 8 \\
        \addlinespace
        \multirow[c]{4}{*}{\textbf{21.0 B}} & 1 & 8 & 40,000 & 32 \\
        & 4 & 4 & 40,000 & 16 \\
        & 8 & 4 & 40,000 & 8 \\
        & 16 & 2 & 40,000 & 8 \\
        \bottomrule
    \end{tabular} }
    \caption{Overview of the total compute budget and resources used for different \textsc{x-elm} experiments. \textbf{k} is the number of experts, \textbf{\# GPUs} indicates the number of GPUs used to train each expert, and \textbf{grad acc.} gives the number of gradient accumulation steps used.}
    \label{tab:app-resources}
\end{table}

\section{Additional Experimental Details}

\subsection{Pretraining}
\label{app:params}
Table \ref{tab:app-data} summarizes the languages we use, as well as their frequencies in the original XGLM pretraining dataset and in our sub-sampled mC4 corpus. 

Table \ref{tab:app-resources} presents the compute allocated to each expert and setting at different compute budgets of the \textsc{x-elm} experiments. The per-model instance batch size (\textbf{bsz}) for all experiments is 2, and each training example had a sequence length (\textbf{seq. len}) of 2048. The total token budget (\textbf{\# Tokens}) is the product of ($k$, \# GPUs, \# updates, grad acc., bsz, seq. len), normalized by the number of GPUs used for model parallelism (2). 

The experts are trained with a linear decay learning rate schedule; we use a maximum learning rate of $1.5e-4$ after performing preliminary learning rate sweeps. 

\subsection{In-Context Learning}
\label{app:icl}
We reimplement the evaluation protocol from \citet{lin2022fewshot}, where the model scores multiple versions of every example (with the different possible labels filled in), and the label of the highest-scoring version is considered as the model's prediction. We use the English prompt formats and evaluation protocols developed for the seed LM of our experts, XGLM, for the downstream tasks of XNLI, XStoryCloze, and PAWS-X. The prompt templates we use are reprorted in Table \ref{tab:app-prompts}. 

In the few-shot setting, we perform five evaluation runs with different demonstration samples and reported the average performance. All few-shot experiments are performed with eight random demonstrations. Unless otherwise stated, we evaluate performance on the development set and sample demonstrations from the training set. As we are testing \textsc{x-elm}'s cross-lingual abilities, the demonstrations are in English for every target language.

\subsection{Licensing and Intended Use} All of the artifacts used to build the \textsc{x-elm}s presented in this work were released for use within academic research \cite{gururangan2023scaling, lin2022fewshot, xue2021mt5}. This also holds for the evaluation benchmarks used to validate the \textsc{x-elm}s \cite{xue2021mt5, conneau2018xnli, lin2022fewshot, yang2019paws}. Therefore, the intended use of the code and models presented in this work is in open-source research as well; paricularly, to improve the performance and useablilty of multilingual models for all languages.

\begin{table}[H]
    \centering
    \small
    \resizebox{0.99\linewidth}{!}{ \begin{tabular}{l | r r }
        \toprule
         \textbf{Language} & \textbf{mC4$^\dagger$ Size (\%)} & \textbf{XGLM Size} \\
         AR (Arabic) & 243.14 (4.1\%) & 64.34 \\
         BG (Bulgarian) & 109.3 (1.9\%) & 61.10 \\
         DE (German) & 615.59 (10.4\%) & 369.30 \\
         EL (Greek) & 193.63 (3.3\%) & 180.37 \\
         EN (English) & 877.43 (14.8\%) & 3,324.45 \\
         ES (Spanish) & 723.17 (12.2\%) & 363.83 \\
         FR (French) & 506.74 (8.6\%) & 303.76 \\
         HI (Hindi) & 125.44 (2.1\%) & 26.63 \\
         JA (Japanese) & 764.71 (12.9\%) & 293.39 \\
         KO (Korean) & 91.29 (1.5\%) & 79.08 \\
         RU (Russian) & 957.02 (16.2\%) & 1,007.38 \\
         SW (Swahili) & 3.06 (0.05\%) & 3.19 \\
         TR (Turkish) & 248.07 (4.2\%) & 51.51 \\
         UR (Urdu) & 10.15 (0.2\%) & 7.77 \\
         VI (Vietnamese) & 296.65 (5.0\%) & 50.45 \\
         ZH (Chinese) & 143.68 (2.4\%) & 485.32 \\
         \addlinespace
         AZ (Azerbaijani) & 15.23 (--) & -- \\
         HE (Hebrew) & 67.14 (--) & -- \\
         PL (Polish) & 393.85 (--) & -- \\
         SV (Swedish) & 154.54 (--) & -- \\
        \bottomrule
    \end{tabular} }
    \caption{The frequencies and relative percentages of different languages in our training corpus ($^\dagger$an mC4 subsample) and in the XGLM corpus, CC100-XL \cite{lin2022fewshot}. Sizes are in gigabytes (GiB). EN, ES, FR, and RU are downsampled to 1,024 shards for mC4.}
    \label{tab:app-data}
\end{table}

\begin{table*}[t]
    \centering
    \small
    \resizebox{0.99\linewidth}{!}{ \begin{tabular}{c|c|c}
        \toprule
        \textbf{Dataset} & \textbf{Prompt} & \textbf{Labels} \\
        XNLI & \texttt{\{Sentence 1\}, right? [Mask], \{Sentence 2\}} & Entailment: \texttt{Yes} | Neural: \texttt{Also} | Contradiction: \texttt{No} \\
        XStoryCloze & \texttt{\{Context\} [Mask]} & Identity \\
        PAWS-X & \texttt{\{Sentence 1\}, right? [Mask], \{Sentence 2}\} & True: \texttt{Yes} | False: \texttt{No} \\
        \bottomrule
    \end{tabular} }
    \caption{Prompts used for the ICL experiments in \S \ref{sec:icl-results}; the \texttt{[MASK]} is filled with one of the label forms given in the last column. For XStoryCloze, \texttt{\{Context\}} refers to the format \texttt{\{Sent. 1\} \{Sent. 2\} \{Sent. 3\} \{Sent. 4\}}, and ``Identity'' refers to the text of one of the answers given for that example.}
    \label{tab:app-prompts}
\end{table*}

\section{Additional \textsc{x-elm} Analysis}

\subsection{Comparing the Data Distribution of Clustering Techniques}

\begin{figure}
    \centering
    \includegraphics[width=\linewidth]{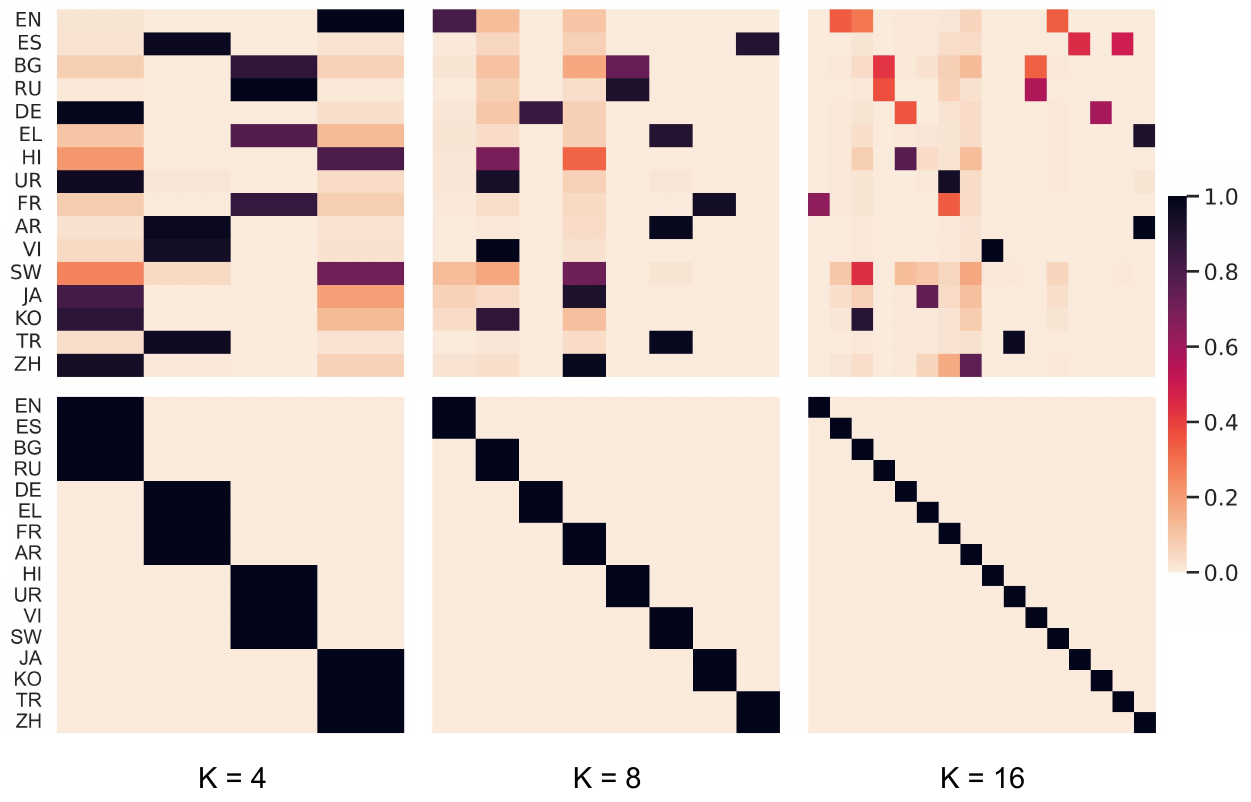}
    \caption{Percentage of language data assigned to different experts with TF-IDF (top row) and Typ. (bottom row) clustering. For Typ. clustering, each language is assigned entirely to a single expert.}
    \label{fig:cluster-dist}
\end{figure}

Figure \ref{fig:cluster-dist} shows the difference in language distributions between the \textit{TF-IDF} and \textit{Linguistic Typology} clusters. While \textit{TF-IDF} allows language data to spread across experts, we find that, in practice, the distributions remain relatively sparse. The main exception is at $k=16$, when the highest-resourced languages in the data (e.g., English or Russian) are split across clusters due to the constraint that balances the amount of data per cluster.

\subsection{Sparse TF-IDF Ensembling}
In \S \ref{sec:results}, we compare ensembling TF-IDF experts in an \textsc{x-elm} set against choosing a single TF-IDF expert for inference based on the amount of in-language data seen by that expert during training. In the cases of $m$=2,4, this approach sparsifies the ensemble by dynamically selecting the top $m$ experts based on their current ensemble weights. Here, we additionally consider how \textit{sparsifying} the TF-IDF ensemble holds up against these other settings (Table \ref{tab:app-sparse-tfidf}). We find that for seen languages, reducing the number of experts active to just $m$=2 usually gives very similar performance to the full ensemble ($m$=8). However, this is not true in the case of \textit{unseen} languages, where the $m$=8 setting consistently outperforms sparser ensembles. 

\subsection{\textsc{x-elm} Forgetting}
\label{sec:results-forgetting}
We evaluate \textsc{x-elm}s as a set of models by dynamically choosing the best expert for a given evaluation setting or ensembling the experts' outputs. However, each expert is initialized with a model trained on all the languages we consider. This prompts the question: how much do individual experts \textit{forget}\footnote{We consider an expert to have \textit{forgetten} information about a language if its perplexity on that language increases.} about the languages they are not specialized to?

\begin{figure}[b!]
    \centering
    \begin{minipage}{.49\textwidth}
    \includegraphics[width=\linewidth]{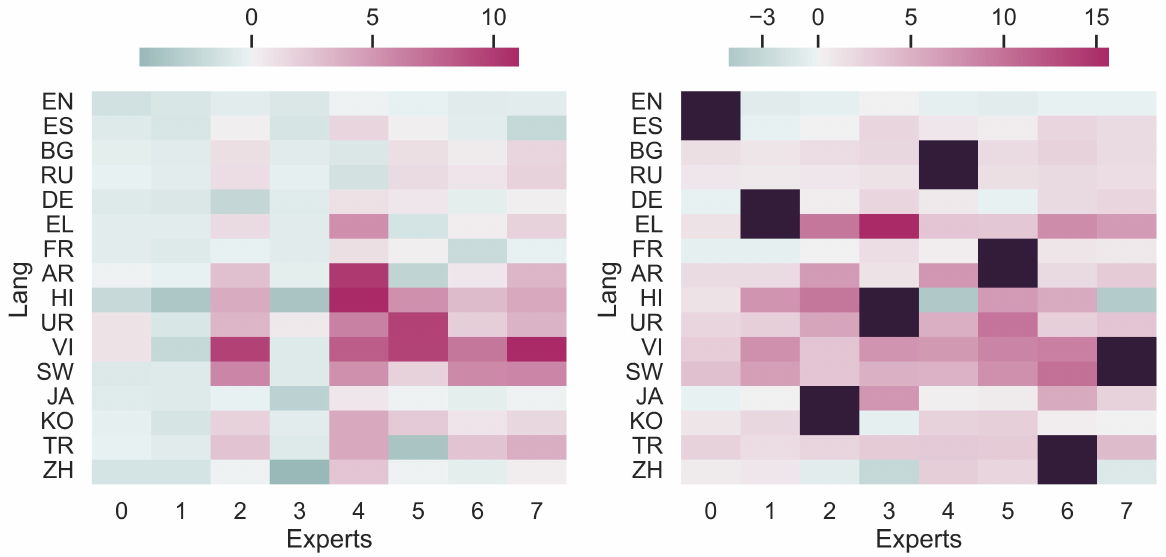}
    \end{minipage} %
    \begin{minipage}{.49\textwidth}
    \includegraphics[width=\linewidth]{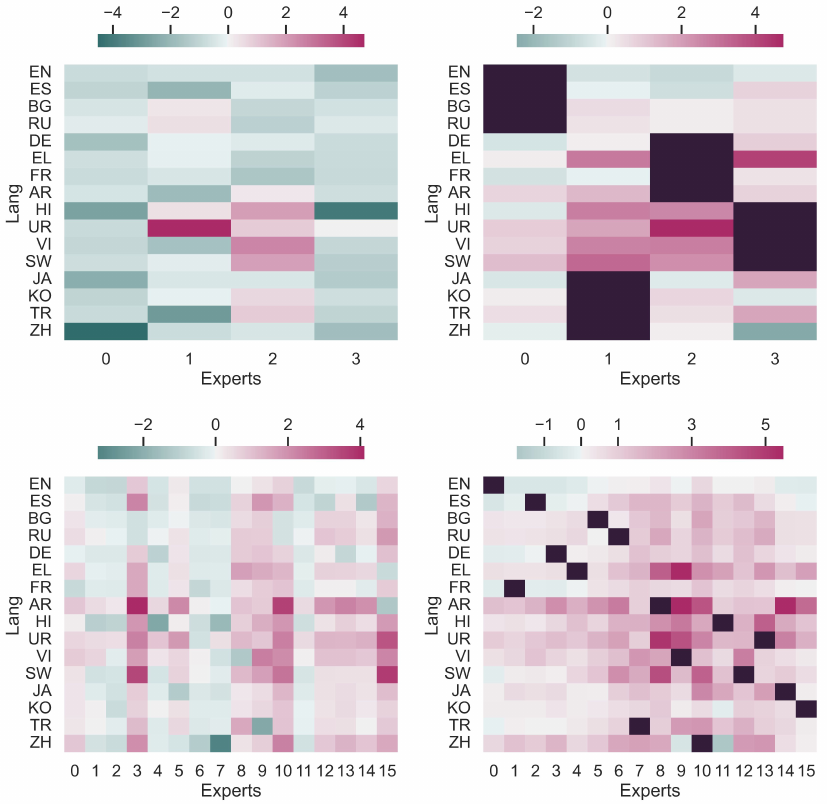}
    \end{minipage}
    \caption{Heatmap comparing individual \textsc{x-elm} perplexities to the seed LM with TF-IDF (left) and Typ. (right). Rows give results for $k=8, 4, 16$, respectively. \textcolor{custompink}{Positive} scores indicate that the expert \textit{forgot} that language. For Typ. clusters, languages that the model was explicitly trained on are grayed out.}
    \label{fig:app-forgetting}
\end{figure}

\paragraph{Forgetting occurs as \textsc{x-elm}s become more specialized.} We compare the perplexity of each expert model on all pretraining languages to that of the seed model, XGLM-1.7B (Figure \ref{fig:app-forgetting}).%
Across the considered values of $k$, we see less forgetting in the \textsc{x-elm}s trained on TF-IDF clusters than in those clustered typologically. For the $k=8$ expert setting, the TF-IDF experts only forget on 47.7\% of settings, and when forgetting occurs, the perplexity increase over the baseline is 3.10 on average. For typologically clustered experts, these measures are 83.6\% and 3.14, respectively. %

We observe similar trends for the $k=4$ and $k=16$ \textsc{x-elm}s. On average, the $k=4$ TF-IDF experts experience forgetting in only 18.8\% of cases with an average perplexity increase of 1.24 when forgetting occurs; the typology experts forget 78.1\% of the time with an average perplexity increase of 1.34. For the $k=16$ setting, these statistics are 60.9\% and 0.9 for the TF-IDF clusters and 89.4\% and 1.24 for the typology clusters.

\begin{figure}[]
    \centering
    \includegraphics[width=0.99\linewidth]{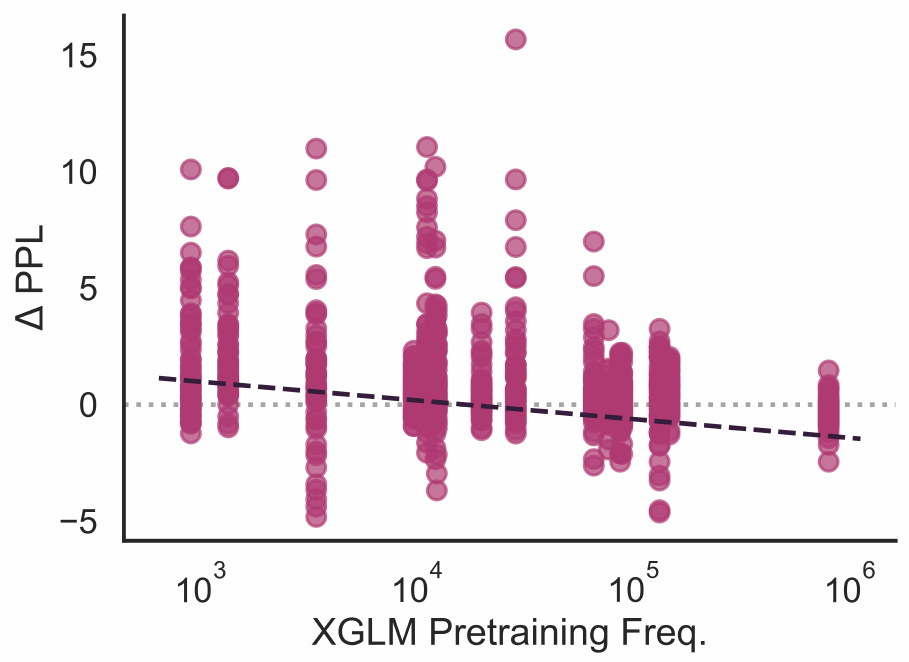}
    \caption{Per-expert deltas compared to the original XGLM-1.7B of every pretraining language plotted against the language's frequency in the original XGLM pretraining corpus ($\rho = -0.33$, $p<<0.001$).}
    \label{fig:forgetting-scatter}
\end{figure} 

\begin{table*}[]
    \centering
    \small
    \begin{tabular}{c | r r | r r r | r r r } %
        \toprule
        \multirow{2}{*}{\textbf{Lang.}} & & & \multicolumn{3}{|c}{\textbf{k=4 Experts}} & \multicolumn{3}{|c}{\textbf{k=16 Experts}} \\
        & XGLM & Dense & TF-IDF$_{top1}$ & TF-IDF$_{ens}$$^*$ & Typ. & TF-IDF$_{top1}$ & TF-IDF$_{ens}$$^*$ & Typ. \\
        \addlinespace
        \textbf{AR} & 16.85 & 15.29 & 14.99 & 15.03 & 15.00 & 15.60 & 15.67 & 15.40 \\
        \textbf{BG} & 11.31 & 10.44 & 10.39 & 10.39 & 10.42 & 11.10 & 10.70 & 10.31 \\
        \textbf{DE} & 15.53 & 14.02 & 13.85 & 13.89 & 13.71 & 14.71 & 14.43 & 14.5 \\
        \textbf{EL} & 10.44 & 9.40 & 9.36 & 9.33 & 9.28 & 9.72 & 9.64 & 9.41 \\
        \textbf{EN} & 14.37 & 12.88 & 12.64 & 12.71 & 12.78 & 13.60 & 13.23 & 13.27 \\
        \textbf{ES} & 16.02 & 14.13 & 13.93 & 13.96 & 14.06 & 14.83 & 14.58 & 14.59 \\
        \textbf{FR} & 13.12 & 11.78 & 11.62 & 11.65 & 11.55 & 12.38 & 12.13 & 12.15 \\
        \textbf{HI} & 18.28 & 14.28 & 14.22 & 14.21 & 12.64 & 16.11 & 15.67 & 13.86 \\
        \textbf{JA} & 14.57 & 12.31 & 12.23 & 12.12 & 11.73 & 13.39 & 13.14 & 13.18 \\
        \textbf{KO} & 8.82 & 7.78 & 7.81 & 7.77 & 7.70 & 8.14 & 8.09 & 7.75 \\
        \textbf{RU} & 13.43 & 12.52 & 12.30 & 12.33 & 12.46 & 12.96 & 12.76 & 12.82 \\
        \textbf{SW} & 19.85 & 18.70 & 18.61 & 18.62 & 18.19 & 19.38 & 19.13 & 16.43 \\
        \textbf{TR} & 17.81 & 15.34 & 14.85 & 14.96 & 14.81 & 15.67 & 15.78 & 15.52 \\
        \textbf{UR} & 14.38 & 13.45 & 13.56 & 13.73 & 13.18 & 13.88 & 13.87 & 12.65 \\
        \textbf{VI} & 13.07 & 11.39 & 11.43 & 11.21 & 10.32 & 11.85 & 11.65 & 11.59 \\
        \textbf{ZH} & 17.91 & 13.74 & 13.38 & 13.70 & 13.11 & 14.65 & 14.95 & 13.58 \\
        \addlinespace
        \textbf{Avg.} & 14.74 & 12.97 & 12.82 & 12.85 & 12.56 & 13.62 & 13.46 & 12.94 \\
        \bottomrule
        
    \end{tabular}
    \caption{Per-language and average perplexity results for the $k=4$ and $k=16$ \textsc{x-elm} experiments (original XGLM and $k=1$ dense model included for comparison). Lower numbers are better. Each \textsc{x-elm} setting is trained on 10.5B tokens.  $^*$TF-IDF ensemble uses more parameters for inference than other evaluations.} %
    \label{tab:app-ppl-results}
\end{table*}
 
\paragraph{\textsc{x-elm}s are more likely to forget certain languages.} For example, English is rarely forgotten, with only 25\% of experts performing worse than the baseline. In comparison, 94.6\% of experts perform worse on Urdu than XGLM. One potential cause of this discrepancy is the frequency with which the language was seen during seed training: languages that are more common in the XGLM pretraining corpus see fewer cases of forgetting and have smaller perplexity increases when it does occur (Figure \ref{fig:forgetting-scatter}). Another likely factor is inaccurate language classification in the BTM training data, which is a common issue when training language models on specific languages \cite{blevins2022language}; this could lead to related, higher-resourced languages contaminating the datasets for lower-resourced ones \cite{kreutzer2022quality}.

\section{Full Experimental Results}
\label{app:full-results}
Table \ref{tab:app-ppl-results} presents the full perplexity results for the $k=4$ and $k=16$ \textsc{x-elm} experiments, trained on a 10.5B token compute budget. We find that both choices of $k$ underperform the $k=8$ setting.

\paragraph{Downstream Evaluation on Individual Languages}
Tables \ref{tab:downstream-xnli}, \ref{tab:downstream-xstorycloze}, and \ref{tab:downstream-pawsx} detail the per-language results for XNLI, XStoryCloze, and PAWS-X, respectively. 

\begin{table}
    \centering
    \small
    \begin{tabular}{c|r r r r}
        \toprule
         \multirow{2}{*}{\textbf{Lang.}} & &  \multicolumn{3}{c}{\textbf{TF-IDF Ens.}} \\
         & top-1 & $m$=2 & $m$=4 & $m$=8 \\
        \addlinespace
        \textbf{AR} & 14.00 & 14.12 & 14.05 & 14.05 \\
        \textbf{BG} & 10.27 & 10.27 & 10.27 & 10.27 \\
        \textbf{DE} & 12.95 & 13.09 & 13.07 & 13.04 \\
        \textbf{EL} & 9.03  & 9.03  & 8.99  & 9.00 \\
        \textbf{EN} & 12.68 & 12.50 & 12.48 & 12.47 \\
        \textbf{ES} & 13.54 & 13.40 & 13.39 & 13.37 \\
        \textbf{FR} & 10.79 & 10.92 & 10.88 & 10.88 \\
        \textbf{HI} & 14.36 & 13.47 & 13.62 & 13.62 \\
        \textbf{JA} & 11.36 & 11.35 & 11.37 & 11.37 \\
        \textbf{KO} & 7.61  & 7.53  & 7.53  & 7.53 \\
        \textbf{RU} & 11.83 & 11.90 & 11.90 & 11.90 \\
        \textbf{SW} & 19.04 & 18.67 & 18.67 & 18.67 \\
        \textbf{TR} & 13.41 & 13.58 & 13.58 & 13.58 \\
        \textbf{UR} & 13.26 & 13.52 & 13.52 & 13.52 \\
        \textbf{VI} & 10.56 & 10.41 & 10.41 & 10.42 \\
        \textbf{ZH} & 12.61 & 12.84 & 12.84 & 12.87 \\
        \addlinespace
        \textbf{Avg.} & 12.33 & 12.29 & 12.29 & 12.28 \\
        \addlinespace
        \textbf{AZ} & -- & 736.49 & 724.97 & 722.10 \\
        \textbf{HE} & -- & 749.12 & 719.68 & 719.05 \\
        \textbf{PL} & -- & 177.31 & 175.27 & 174.83 \\
        \textbf{SV} & -- & 95.33 & 94.37 & 94.14 \\
         \bottomrule
    \end{tabular}
    \caption{Perplexity scores of the different inference methods on the TF-IDF \textsc{x-elm}s trained with 21B tokens. \textbf{Top-1} chooses a single expert per language, with no routing mechanism, whereas \textbf{m=2,4,8} ensembles TF-IDF experts.}
    \label{tab:app-sparse-tfidf}
\end{table}

\begin{table*}[]
    \centering
    \small
    \resizebox{0.99\linewidth}{!}{ \begin{tabular}{l| r r r r r r r r r r r r r r r}
    \toprule
      \textbf{Model} & \textbf{AR} & \textbf{BG} & \textbf{DE} & \textbf{EL} &	\textbf{EN} & \textbf{ES} &	\textbf{FR} & \textbf{HI} & \textbf{RU} & \textbf{SW} & \textbf{TH$^*$} & \textbf{TR} & \textbf{UR} & \textbf{VI} & \textbf{ZH} \\
       \addlinespace
       \multicolumn{2}{l}{\textbf{Zero-shot}} \\
        XGLM (1.7B) & 46.8 & 45.7 & 44.1 & 42.5 & 51.5 & 36.5 & 47.2 & 45.9 & 47.3 & 43.6 & 44.9 & 42.5 & 43.5 & 43.9 & 46.9 \\
        Dense & 47.9 & 45.0 & 45.3 & 45.2 & 51.1 & 37.2 & 45.9 & 44.5 & 44.5 & 39.6 & 44.3 & 44.8 & 43.1 & 41.6 & 44.6 \\
        Typ. (TRG) & 46.2 & 44.9 & 43.9 & 45.4 & 52.0 & 36.0 & 47.2 & 43.5 & 41.9 & 40.6 & -- & 44.2 & 41.9 & 44.4 & 46.3 \\
        TF-IDF (Top-1) & 47.3 & 45.1 & 42.9 & 47.1 & 51.5 & 36.3 & 45.6 & 43.1 & 40.6 & 38.7 & -- & 45.0 & 43.2 & 41.8 & 44.6 \\ %
        TF-IDF (Ens.) & 48.6 & 47.2 & 46.2 & 43.1 & 53.0 & 37.0 & 47.5 & 45.7 & 45.6 & 40.0 & 45.8 & 44.1 & 44.2 & 42.6 & 46.6 \\
       \addlinespace
       \multicolumn{2}{l}{\textbf{Few-shot}} \\
       XGLM (1.7B) & 42.0 & 44.2 & 43.4 & 43.4 & 47.2 & 38.1 & 45.5 & 40.4 & 43.1 & 41.4 & 41.9 & 38.0 & 39.7 & 42.2 & 44.3 \\
       Dense & 43.4 & 42.2 & 43.6 & 41.9 & 45.9 & 36.7 & 42.3 & 42.2 & 40.8 & 40.0 & 43.2 & 39.9 & 40.3 & 41.0 & 43.5 \\
       Typ. (TRG) & 42.8 & 43.0 & 42.6 & 43.0 & 47.3 & 38.5 & 45.4 & 38.9 & 39.9 & 41.7 & -- & 41.0 & 39.6 & 42.9 & 43.4 \\
       Typ. (EN) & 42.2 & 42.6 & 44.0 & 42.6 & 47.3 & 38.5 & 42.9 & 42.1 & 42.8 & 40.9 & 44.5 & 41.1 & 40.0 & 42.1 & 44.9 \\
       TF-IDF (Top-1) & 43.1 & 43.6 & 43.2 & 41.7 & 47.5 & 38.2 & 45.3 & 42.1 & 40.5 & 41.9 & -- & 41.1 & 41.4 & 42.1 & 44.1 \\
       TF-IDF (Ens.) & 43.0 & 43.3 & 44.3 & 43.3 & 47.8 & 37.7 & 44.2 & 43.2 & 42.3 & 41.4 & 44.4 & 41.8 & 41.0 & 42.7 & 44.9 \\ 
    \bottomrule
    \end{tabular} }
    \caption{Individual language accuracy on XNLI. $^*$TH (Thai) is an unseen language for the \textsc{x-elm} models.}
    \label{tab:downstream-xnli}
\end{table*}

\begin{table*}[]
    \centering
    \small
    \begin{tabular}{l| r r r r r r r r r r r}
    \toprule
      \textbf{Model} & \textbf{AR} & \textbf{EN} & \textbf{ES} & \textbf{EU$^{*}$} & \textbf{HI} & \textbf{ID$^{*}$} & \textbf{MY$^{*}$} & \textbf{RU} & \textbf{SW} & \textbf{TE$^{*}$} & \textbf{ZH} \\
       \addlinespace
       \multicolumn{2}{l}{\textbf{Zero-shot}} \\
       XGLM (1.7B) & 53.3 & 63.1 & 57.3 & 56.4 & 55.0 & 59.3 & 54.0 & 60.0 & 60.1 & 57.0 & 55.5 \\
        Dense & 50.5 & 60.7 & 56.1 & 52.1 & 52.0 & 55.4 & 53.4 & 58.6 & 58.6 & 55.5 & 56.2 \\
        Typ. (TRG) & 52.3 & 62.7 & 57.5 & -- & 52.7 & -- & -- & 60.2 & 60.3 & -- & 58.8 \\
        TF-IDF (Top-1) & 52.1 & 62.1 & 58.1 & 53.2 & 55.2 & 57.7 & 52.6 & 59.6 & 60.5 & 57.3 & 57.0 \\
        TF-IDF (Ens.) & 51.9 & 60.4 & 57.8 & 54.0 & 55.4 & 58.5 & 52.0 & 59.5 & 60.2 & 57.1 & 57.0 \\
       \addlinespace
       \multicolumn{2}{l}{\textbf{Few-shot}} \\
        XGLM (1.7B) & 48.6 & 58.2 & 53.2 & 51.7 & 50.4 & 52.1 & 51.5 & 52.5 & 56.0 & 56.5 & 53.7 \\
        Dense & 50.2 & 59.0 & 54.6 & 51.3 & 51.6 & 53.5 & 52.9 & 56.9 & 57.8 & 54.2 & 55.2 \\
       Typ. (TRG) & 50.3 & 60.1 & 55.0 & -- & 52.0 & -- & -- & 57.4 & 58.0 & -- & 56.0 \\
        Typ. (EN) & 48.8 & 60.1 & 55.0 & -- & 52.2 & -- & -- & 53.7 & 57.4 & -- & 55.2 \\
       TF-IDF (Top-1) & 49.3 & 59.5 & 54.5 & 51.4 & 52.4 & 55.2 & 52.9 & 55.4 & 58.0 & 56.1 & 56.1 \\
       TF-IDF (Ens.) & 49.4 & 59.0 & 53.8 & 51.1 & 52.5 & 54.5 & 52.0 & 55.1 & 57.8 & 55.0 & 55.4 \\
    \bottomrule
    \end{tabular}
    \caption{Individual language accuracy on XStoryCloze (and EN StoryCloze). $^*$Unseen languages for the \textsc{x-elm} models.}
    \label{tab:downstream-xstorycloze}
\end{table*}

\begin{table*}[]
    \centering
    \small
    \begin{tabular}{l| r r r r r r r }
    \toprule
      \textbf{Model} & \textbf{DE} & \textbf{EN} & \textbf{ES} & \textbf{FR} & \textbf{JA} & \textbf{KO} & \textbf{ZH} \\
       \addlinespace
       \multicolumn{2}{l}{\textbf{Zero-shot}} \\
       XGLM (1.7B) & 44.5 & 47.9 & 51.8 & 45.2 & 53.8 & 49.6 & 47.0\\
       Dense & 49.4 & 47.5 & 50.7 & 47.5 & 48.8 & 47.2 & 48.0 \\
       Typ. (TRG) & 47.9 & 47.9 & 53.0 & 45.5 & 55.4 & 53.6 & 45.7 \\
       TF-IDF (Top-1) & 47.4 & 46.9 & 55.0 & 45.9 & 54.9 & 49.4 & 50.8 \\
       TF-IDF (Ens.) & 49.1 & 47.1 & 52.1 & 47.2 & 53.6 & 50.0 & 50.4 \\
       \addlinespace
       \multicolumn{2}{l}{\textbf{Few-shot}} \\
       XGLM (1.7B) & 56.3 & 50.5 & 55.4 & 55.2 & 55.6 & 53.0 & 55.7 \\
       Dense & 56.0 & 54.9 & 55.8 & 55.2 & 54.9 & 53.8 & 53.0 \\
       Typ. (TRG) & 56.5 & 53.4 & 55.8 & 55.1 & 55.6 & 55.9 & 55.4 \\
       Typ. (EN) & 56.0 & 53.4 & 55.8 & 55.4 & 55.5 & 54.7 & 55.1 \\
       TF-IDF (Top-1) & 56.6 & 54.2 & 55.7 & 54.9 & 55.6 & 55.7 & 55.7 \\
       TF-IDF (Ens.) & 53.8 & 54.9 & 54.8 & 53.6 & 55.3 & 55.2 & 54.4 \\ 
    \bottomrule
    \end{tabular}
   \caption{Individual language accuracy on PAWS-X.}
    \label{tab:downstream-pawsx}
\end{table*}

\end{document}